# Health, Psychosocial, and Social issues emanating from COVID-19 pandemic based on Social Media Comments using Natural Language Processing

Oladapo Oyebode[1], Chinenye Ndulue[1], Ashfaq Adib[1], Dinesh Mulchandani[1], Banuchitra Suruliraj[1], Fidelia Anulika Orji[4], Christine Chambers[2], Sandra Meier[3], and Rita Orji[1]

[1] Faculty of Computer Science, Dalhousie University, Halifax, Nova Scotia, Canada
[2] Department of Psychology and Neuroscience, Dalhousie University, Halifax, Nova Scotia, Canada
[3] Department of Psychiatry, Faculty of Medicine, Dalhousie University, Halifax, Nova Scotia, Canada
[4] Department of Computer Science, University of Saskatchewan, Saskatoon, Saskatchewan, Canada

**Corresponding Author:**
Oladapo Oyebode
Faculty of Computer Science
Dalhousie University
Halifax, Nova Scotia, B3H 4R2
Canada
Email: oladapo.oyebode@dal.ca

## Abstract

**Background:** The COVID-19 pandemic has caused a global health crisis that affects many aspects of human lives. In the absence of vaccines and antivirals, several behavioural change and policy initiatives, such as physical distancing, have been implemented to control the spread of the coronavirus. Social media data can reveal public perceptions toward how governments and health agencies across the globe are handling the pandemic, as well as the impact of the disease on people regardless of their geographic locations in line with various factors that hinder or facilitate the efforts to control the spread of the pandemic globally.

**Objective:** This paper aims to investigate the impact of the COVID-19 pandemic on people globally using social media data.

**Methods:** We apply natural language processing (NLP) and thematic analysis to understand public opinions, experiences, and issues with respect to the COVID-19 pandemic using social media data. First, we collect over 47 million COVID-19-related comments from Twitter, Facebook, YouTube, and three online discussion forums. Second, we perform data preprocessing which involves applying NLP techniques to clean and prepare the data for automated theme extraction. Third, we apply context-aware NLP approach to extract meaningful keyphrases or themes from over 1 million randomly-selected comments, as well as compute sentiment scores for each theme and assign sentiment polarity (i.e., positive, negative, or neutral) based on the scores using lexicon-based technique. Fourth, we categorize related themes into broader themes.

**Results:** A total of 34 negative themes emerged, out of which 15 are health-related issues, psychosocial issues, and social issues related to the COVID-19 pandemic from the public perspective. Some of the health-related issues are *increased mortality*, *health concerns*, *struggling health systems*, and *fitness issues*; while some of the psychosocial issues include *frustrations due to life disruptions*, *panic shopping*, and *expression of fear*. Social issues include *harassment*, *domestic violence*, and *wrong societal attitude*. In addition, 20 positive themes emerged from our results. Some of the positive themes include *public awareness*, *encouragement*, *gratitude*, *cleaner environment*, *online learning*, *charity*, *spiritual support*, and *innovative research*.

**Conclusions:** We uncover various negative and positive themes representing public perceptions toward the COVID-19 pandemic and recommend interventions that can help address the health, psychosocial, and social issues based on the positive themes and other remedial ideas rooted in research. These interventions will help governments, health professionals and agencies, institutions, and individuals in their efforts to curb the spread of COVID-19 and minimize its impact, as well as in reacting to any future pandemics.

## Introduction

### Background

Infectious diseases have occurred in the past and continue to emerge. Infectious diseases are termed "emerging" if they newly appear in a population or have existed but are increasing rapidly in incidence or geographic range [1]. Examples of emerging infectious diseases include acquired immunodeficiency syndrome, Ebola, dengue hemorrhagic fever, Lassa fever, severe acute respiratory syndrome, H1N1 flu, Zika, etc. [2]. Evidence shows that emerging infectious diseases are among the leading causes of death and disability globally [3]. For instance, a 1-year estimate of the 2009 H1N1 flu pandemic shows that 43 to 89 million people were infected [4] and 201,200 respiratory deaths and 83,300 cardiovascular deaths were linked to the disease [5] globally. In addition, 770,000 HIV deaths were recorded in 2018 alone, with approximately 37.9 million people already infected with the virus globally [6]. Ebola is another deadly infectious disease that has an average case fatality rate of about 50%, with a range of 25% to 90% case fatality rates in past outbreaks [2,7].

In December 2019, the novel coronavirus which causes the COVID-19 disease emerged and soon became the latest deadly infectious disease [8,9] globally with more than 9.4 million confirmed cases and over 482,800 deaths in 188 countries/regions as at June 25, 2020 [10]. Hence, it was declared a pandemic by the World Health Organization. The COVID-19 pandemic has strained the global health systems, and also caused socio-economic challenges due to job losses and lockdowns (and other restrictive measures) imposed by governments and public health agencies to curtail the spread of the virus. Evidence has already shown that emerging infectious diseases impose significant burden on global economies and public health [3,11–13]. To understand public concern and personal experiences, and factors that hinder or facilitate the efforts to control the spread of the COVID-19 pandemic globally, social media data can produce rich and useful insights that were previously impossible in both scale and extent [14].

Over the years, social media has witnessed a surge in active users to more than 3.8 billion globally [15], making it a rich source of data for research in diverse domains. In the health domain, social media data (i.e., user comments or posts on Twitter, Facebook, YouTube, Instagram, online forums, blogs, etc.) have been used to investigate mental health issues [16,17], maternal health issues [18,19], diseases [20–24], substance use [25,26] and other health-related issues [27,28]. Other domains (e.g., politics, commerce, marketing, banking) have also witnessed widespread use of social media data to uncover new insights related to election results [29–32], election campaign [33], customer behaviour and engagement [34,35], etc. As regards the COVID-19 crisis, social media data can reveal public perceptions toward how governments and health agencies across the globe are handling the pandemic, as well as the social, economic, psychological, and health-related impact of the disease on people regardless of their geographic locations in line with various factors that hinder or facilitate the efforts to control the spread of the COVID-19 pandemic globally.

In this paper, we apply natural language processing (NLP) to understand public opinions, experiences, and issues with respect to the COVID-19 pandemic using data from Twitter, Facebook, YouTube, and three online discussion forums (i.e., *Archinect.com* [36,37], *LiveScience.com* [38], and *PushSquare.com* [39]). NLP is a well-established method that has been applied in many health informatics papers to understand various health-related issues. For example, Abdalla et al. studied the privacy implications of word embeddings trained on clinical data containing personal health information [40], while Bekhuis et al. applied NLP to extract clinical phrases and keywords from corpus of messages posted to an internet mailing list [41].

The contributions of our research are as follows:

1. We apply a context-aware Natural Language Processing (NLP) approach for extracting opinionated themes from COVID-19-related social media comments.
2. We uncover various negative and positive themes representing public perceptions toward the COVID-19 pandemic. Our results reveal 34 negative themes, out of which 15 are *health-related issues*, *psychosocial issues*, and *social issues* related to the pandemic from the public perspective. In addition, 20 positive themes emerged from our results.

3. We recommend interventions that can help address the health, psychosocial, and social issues based on the positive themes and other remedial ideas rooted in research. These interventions will help governments, health professionals and agencies, institutions, and individuals in their efforts to curb the spread of COVID-19 and minimize its impact, as well as in reacting to any future pandemics.

## Relevant Literature

Social media has been a rich source of data for research in many domains, including health [42]. Research that utilizes social media in conjunction with natural language processing (NLP) within the health domain continues to grow and cover broad application areas such as health surveillance (e.g., mental health, substance use, diseases, pharmacovigilance, etc.), health communication, sentiment analysis, and so on [43]. For example, Park et al. [44] used the lexicon-based approach to track prevalence of keywords indicating public interest in four health issues – Ebola, e-cigarette, marijuana, and influenza – based on social media data. Afterwards, they generated topics or themes that explain changes in discussion volume over time using the Latent Dirichlet Allocation (LDA) algorithm. Similarly, Jelodar et al. [45] applied LDA to extract latent topics in COVID-19-related comments and used the LSTM recurrent neural network technique for sentiment classification. Furthermore, Nobles et al. [46] used social media data to examine the needs (including seeking health information) of reportable sexually transmitted diseases community. Their NLP approach involves extracting top 50 unigrams from the posts based on frequency, and then generating topics or themes using the non-negative matrix factorization technique instead of LDA. Paul et al. [47] applied the Ailment Topic Aspect Model to generate latent topics from Twitter data with the aim of detecting mentions of specific ailments, including allergies, obesity, and insomnia. They used a list of keyphrases to automatically identify possible systems and treatments. McNeill et al. [48] investigated how the dissemination of H1N1-related advice in the United Kingdom encourages/discourages vaccine and antiviral uptake using Twitter data. They conducted an automated content analysis using KH-Coder tool to explore potential topics based on frequency of occurrence, and then performed a more detailed or conversational analysis to understand skepticism over economic beneficiaries of vaccination, as well as the risks and benefits of medication based on public opinion. On the other hand, Oyebode et al. [49] performed sentiment analysis on user reviews of mental health apps using machine learning approach. They compared five classifiers (based on five different machine learning algorithms) and used the best performing classifier to predict the sentiment polarity of reviews. However, none of the approaches above considers the context in which words appear in unstructured texts, which instinctively plays a huge role in conveying meaning.

To investigate the significance of contextual text analysis, Dave et al. [50] compared the non-contextual N-Gram Chunking approach and the contextual Part-of-Speech (POS) chunking approach in their experimental research in the field of advertising. While the N-Gram chunking method simply extracts words of varying lengths within sentence boundary as candidate keyphrases or themes, the POS chunking method infers the context of words using POS patterns such as one or more noun tags (NN, NNP, NNS, and NNPS) along with adjective tags (JJ) and optional cardinal tags (CD) and determiners (DT). They focus on keyphrases of up to length six for their experiments. Their initial assessment showed that majority of the keyphrases generated using the N-Gram chunking method are not meaningful within the advertising context, hence not useful. Furthermore, they observed the impact of keyphrases from both methods on the performance of classification systems based on Naïve Bayes, logistic regression, and bagging machine learning algorithms. Their findings revealed that systems using the POS chunking method outperformed those using the N-Gram chunking method for feature extraction. We leveraged Dave et al.'s contextual method in this work and extended it to capture additional POS patterns, NLP preprocessing techniques, and sentiment scoring using lexicon-based technique.

Finally, to uncover insights about the type of information shared on Twitter during the peak of the H1N1 (swine flu) pandemic in 2009, Ahmed et al. [51] generated eight broad themes using coding method involving expert reviewers. Similarly, Bekhuis et al. [41] involved two dentists to manually and iteratively classify clinical phrases into categories and subcategories. We also used this method in the theme categorization stage of our work to group related themes or keyphrases into broad themes.

## Methods

The main goal of this paper is to understand people's personal experiences and opinions with respect to the COVID-19 pandemic using social media data. To achieve this, we apply various standard and well-known computational techniques which are highlighted below and summarized in Figure 1.

1. We collect COVID-19-related comments or posts from Twitter, Facebook, YouTube, and three online discussion forums using programming languages (Python and C#) and relevant application programming interfaces (APIs).
2. We perform preprocessing tasks which involve applying Natural Language Processing (NLP) techniques to clean the data and prepare them for the theme extraction phase.
3. We apply context-aware NLP approach to extract meaningful themes or keyphrases which are words or phrases that convey the topical content of the comments. This approach is in seven stages: *grammar definition*, *sentence breaking and tokenization*, *part-of-speech tagging*, *lemmatization*, *syntactic parsing*, *transformation and filtering*, and *sentiment scoring*.
4. We categorize related themes into broader themes.

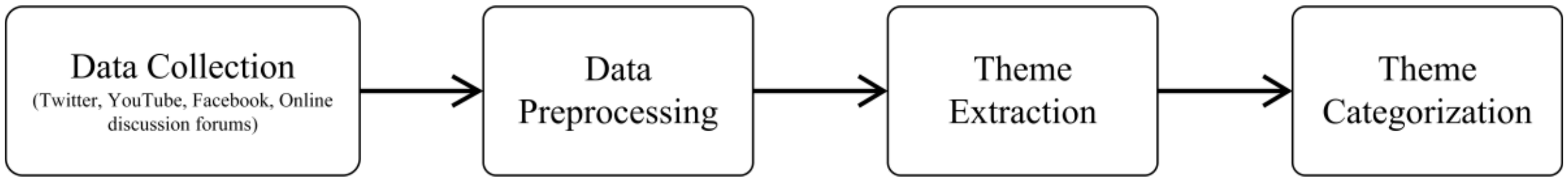

Figure 1. Methodological stages

## Data Collection

We utilized various automated techniques to collect 47,410,795 (over 47 million) COVID-19 or Coronavirus-related comments from six social media platforms – *Twitter*, *YouTube*, *Facebook*, *Archinect.com*, *LiveScience.com*, and *PushSquare.com*. The techniques, as well as the breakdown of the data collected from each platform, are described below:

1. Twitter: We developed a tool using C# programming language to automatically extract tweets containing relevant hashtags in real-time through the Twitter Streaming API [52]. The Twitter hashtags related to COVID-19 include *#QuarantineAndChill*, *#MyPandemicSurvivalPlan*, *#CoronaVirus*, *#caronavirusoutbreak*, *#CoronavirusOutbreak*, *#Quarantined*, *#pandemic*, *#coronapocalypse*, *#COVID19*, *#QuarantineLife*, *#Covid_19*, *#StopTheSpread*, *#CoronaVirusUpdates*, *#StayAtHome*, *#selfquarantine*, *# COVID 一 19*, *#panicbuying*, *#ncov2019*, *#Coronavid19*, *#CoronaCrisis*, *#SocialDistancing*, *#COVID*, *#cronovirus*, *#CoronaVirusUpdate*, *#CoronavirusPandemic*, and *#COVID-19*. A total of 47,249,973 tweets were collected.

2. YouTube: We developed a Python script to retrieve comments associated with relevant videos through the YouTube Data API [53] using the search keywords such as: *covid19*, *covid-19*, and *coronavirus*. Due to YouTube's quota limits, we were only able to extract 111,722 comments across 2,939 videos.

3. Facebook: Due to Facebook's automated search restrictions, we applied a semi-automated approach to extract comments. First, we manually retrieved relevant groups (n=91) and pages (n=68) using the following search keywords: *COVID-19*, *Coronavirus*, and *COVID*. Afterwards, we developed a Python script to extract 777 and 8,382 comments posted on the groups and pages respectively.

4. Online discussion forums: We developed a Python script to extract 20,747; 793; and 18,401 comments (from Coronavirus-related threads) on *Archinect.com*, *LiveScience.com*, and *PushSquare.com* respectively.

## Data Preprocessing

Next, we applied the following NLP techniques to clean and prepare data for analysis using Python:

1. Remove hashtags, mentions, and URLs

2. Expand contractions (e.g., *wouldn't* is replaced with *would not*)

3. Unescape HTML characters (e.g., “&” is replaced with the “&” equivalent)
4. Remove HTML tags (e.g., <p>, <span>, <br />, etc.)
5. Remove special characters, except those with semantic implications such as period and exclamation mark (which are useful for identifying sentence boundary), comma, etc.
6. Reduce repeated characters (e.g., *tooooooool* becomes *tool*)
7. Convert slangs to their equivalent English words using online slang dictionaries [54,55] which contain 5,434 entries in total
8. Remove numeric words

After the preprocessing tasks were completed, non-English and duplicated comments were removed, thereby reducing total number of comments to 8,021,341.

### Theme Extraction

Next, we randomly selected 1,051,616 comments (representing approximately 13% of the entire dataset) and then extracted meaningful keyphrases or themes that convey the topical content of the comments. We refer to the dataset containing the comments as *corpus* and each comment as *document* in the remaining parts of this paper. We focus on themes that are opinionated (i.e., express or imply positive or negative sentiment [56]) since our goal is to determine public opinions and impact with respect to the COVID-19 pandemic. We extracted candidate themes from our corpus using a seven-stage context-aware NLP approach, shown in Figure 2. We implemented our approach using the Python programming language.

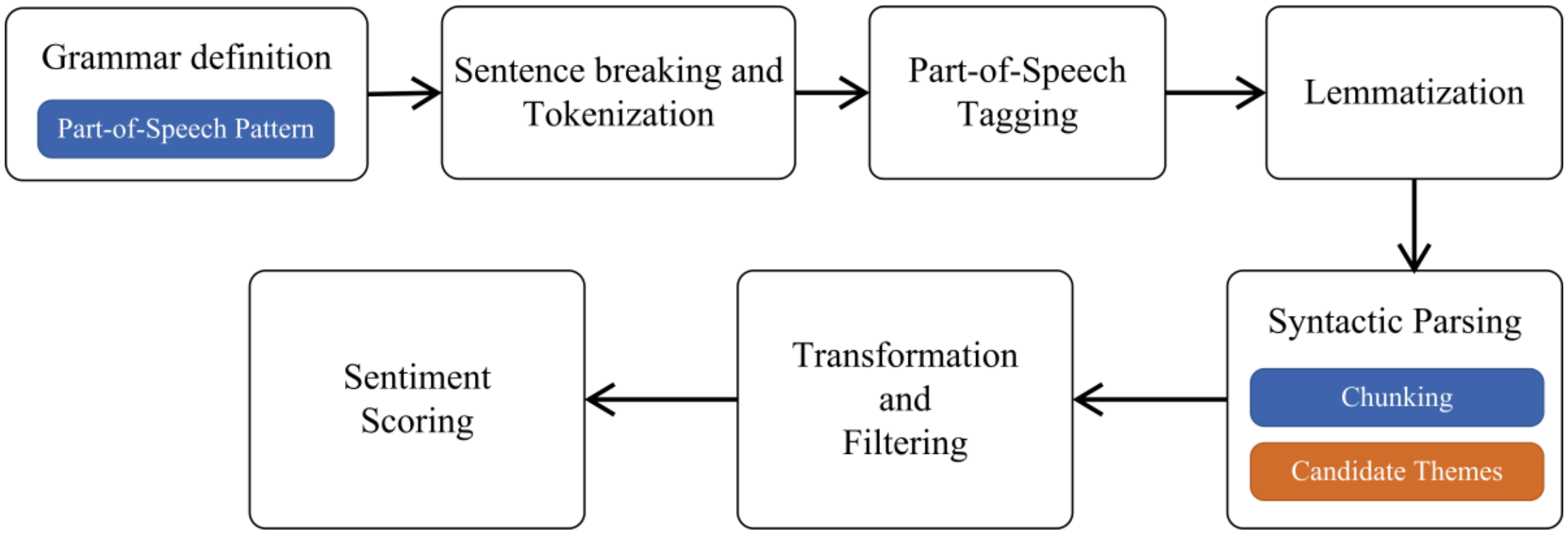

Figure 2. Context-aware Natural Language Processing (NLP) Approach

#### *Grammar definition*

A grammar is a set of rules that describe how syntactic units (such as sentences and phrases) should be deconstructed into their constituents [57]. To derive meaningful themes or keyphrases, we defined a grammar (see below) which specifies a meaningful part-of-speech (POS) pattern that the syntactic parser uses to deconstruct each sentence in the documents. Table 1 shows the various parts of speech (or syntactic categories) covered by our grammar. These syntactic categories are based on well-established part-of-speech tagging guidelines for English [58].

$$\{< \boldsymbol{DT} >? < \boldsymbol{JJ}.*>* < \boldsymbol{NN}.*>* < \boldsymbol{VB}.*>? \ (< \boldsymbol{IN} >? < \boldsymbol{DT} >? < \boldsymbol{JJ}.*>* < \boldsymbol{NN}.*>*)? \ \}$$

In the grammar above, the "?" and "*" characters represent "optional" and "zero or more occurrences" respectively. Our grammar is aimed at generating keyphrases that capture both context and sentiment of a conversation using nouns, adjectives, and verbs. Research has shown that nouns are most useful in knowing the context of a conversation (i.e., what is being discussed) [59], while verbs and adjectives are important for sentiment detection [60]. Determiners and prepositions are also captured by the grammar since they usually co-occur with noun or adjective phrases (e.g., ***a*** *meal* ***for*** *six people*, ***a*** *hospital* ***on the*** *hilltop*, etc.).

Table 1. POS tags, description, and the corresponding matching POS pattern

| **Tag** | **Description** | **Matching Pattern** |
|---|---|---|
| DT | Determiner | <DT> |
| JJ | Adjective | |
| JJR | Adjective (comparative) | <JJ.*> |
| JJS | Adjective (superlative) | |
| NN | Noun (singular) | |
| NNS | Noun (plural) | <NN.*> |
| NNP | Proper Noun (singular) | |
| NNPS | Proper Noun (plural) | |
| VB | Verb (base form) | |
| VBD | Verb (past tense) | |
| VBG | Verb (gerund or present participle) | <VB.*> |
| VBN | Verb (past participle) | |
| VBP | Verb (non-3rd person singular present) | |
| VBZ | Verb (3rd person singular present) | |
| IN | Preposition or subordinating conjunction | <IN> |

### *Sentence Breaking and Tokenization*

Next, each document is split into sentences, and then each sentence is split into tokens or words. The sentence breaking task is achieved using an unsupervised algorithm that considers abbreviations, collocations, capitalizations, and punctuations to detect sentence boundaries [61].

### *Part-of-Speech (PoS) Tagging*

The tagging module associates each token with its part of speech. The part of speech tags are based on the Penn Treebank tagset [58,62], some of which are shown in Table 1.

### *Lemmatization*

Each token is reduced to its root form, depending on its part of speech. This activity is called *lemmatization*. For example, *worse* and *better* which are both adjectives will become *bad* and *good* respectively. Prior to lemmatization, each token is converted to lowercase. While [63] applied stemming for its tokens, we chose lemmatization over stemming since lemmatization returns root words that are always meaningful and exist in the English dictionary. Stemming, on the other hand, may return root words that have no meaning at all since it merely removes prefixes or suffixes based on rule-based method [64].

### *Syntactic Parsing*

The syntactic parsing module deconstructs each sentence into a parse tree and then creates chunks or phrases based on the grammar or POS pattern defined in the first step. In other words, the parser's chunking process involves matching phrases composed of an optional determiner, zero or more of any time of adjective, zero or more of any type of noun, any type of verb (but optional), as well as an optional component. This component consists of an optional preposition, an optional determiner, zero or more of any type of adjective, and zero or more of any type of noun. The output of this stage is the candidate themes ($C_{themes}$).

### *Transformation and Filtering*

In this stage, themes that are stopwords (i.e., words that are commonly used, such as *the*, *a*, *an*, *with*, *in*, *that*, etc.) are removed from $C_{themes}$ using a predefined list $L_{stopwords}$ compiled from multiple sources such as [65]. Also, a subset of $L_{stopwords}$ are removed from the start and end of (and from within) the remaining themes in $C_{themes}$ such that the meaning of the themes is preserved. Afterwards, duplicates were removed from $C_{themes}$. In addition, themes containing more than ten words are removed from $C_{themes}$. While previous research excluded themes above length six [50], we excluded themes above length ten to prevent losing important themes that would have enriched insights from this paper. Since our focus is on opinionated themes (i.e., positive and negative themes), we applied a filtering technique that involves computing sentiment score for each theme and discarding non-opinionated themes.

### *Sentiment Scoring*

To determine opinionated themes in $C_{themes}$, the scoring module computes sentiment score, $S_{score}$, ranging from $-1$ to +1 for each theme using the VADER lexicon-based algorithm [66]. Afterwards, each theme is assigned a polarity based on the sentiment score using the criteria in [66] and shown in Table 2. Themes with the *neutral* polarity are excluded from $C_{themes}$ since they are not opinionated.

Table 2. Criteria for sentiment scoring

| **Criteria** | **Sentiment Polarity** |
|---|---|
| $S_{score} > 0.05$ | Positive |
| $S_{score} < -0.05$ | Negative |
| $S_{score}$ between $-0.05$ and +0.05 | Neutral |

### Theme Categorization

Next, we recruited four reviewers to categorize related themes into a broader theme using thematic analysis method. We assigned the negative themes to a group of two reviewers (G1) and the positive themes to a second group of two reviewers (G2). Each reviewer independently examines the themes iteratively and continues to categorize related themes until saturation level is reached (i.e., no new categories were emerging from the themes). We measured interrater reliability using the percentage agreement metric [67]. The percentage interrater reliability score for G1 is 98.0% while the score for G2 is 99.3%.

## Results

In this section, we discuss the results of our experiments and theme categorization. From the large corpus used for the experiment, a total of 427,875 unique negative themes and 520,685 unique positive themes were automatically generated.

### Negative Themes

Figure 3 shows top 130 sample negative themes and their dominance in terms of frequency of occurrence. Our results revealed that *death* (n=10,187) is the dominant theme, followed by *die* (n=7,240), *fight* (n=5,891), *bad* (n=3,808), *kill* (n=3,668), *lose* (n=3,631), *pay* (n=3486), *leave* (n=3,234), *crisis* (n=2,783), *hard* (n=2,720), *worry* (n=2,476), *sick* (n=2,314), *sad* (n=2,129), and so on. Figure 4 shows more negative themes, such as *national health emergency, scary time*, *life suck*, *everyone struggle*, *dangerous lie*, *child die*, *trouble breathe*, *no medicine*, *sick people*, *pay bill*, *horrible virus*, *fear coronavirus*, *extra cautious*, *steal mask*, *family die*, *people in crisis*, *bad leadership*, *in house bore*, *feel horrible*, *total incompetence*, *call virus hoax*, *conspiracy theory ridiculous*, *take no precaution*, *serious lockdown*, *increase in suicide rate*, *people starve*, *lack of preparedness*, *fight menace*, *restriction on travel*, etc.

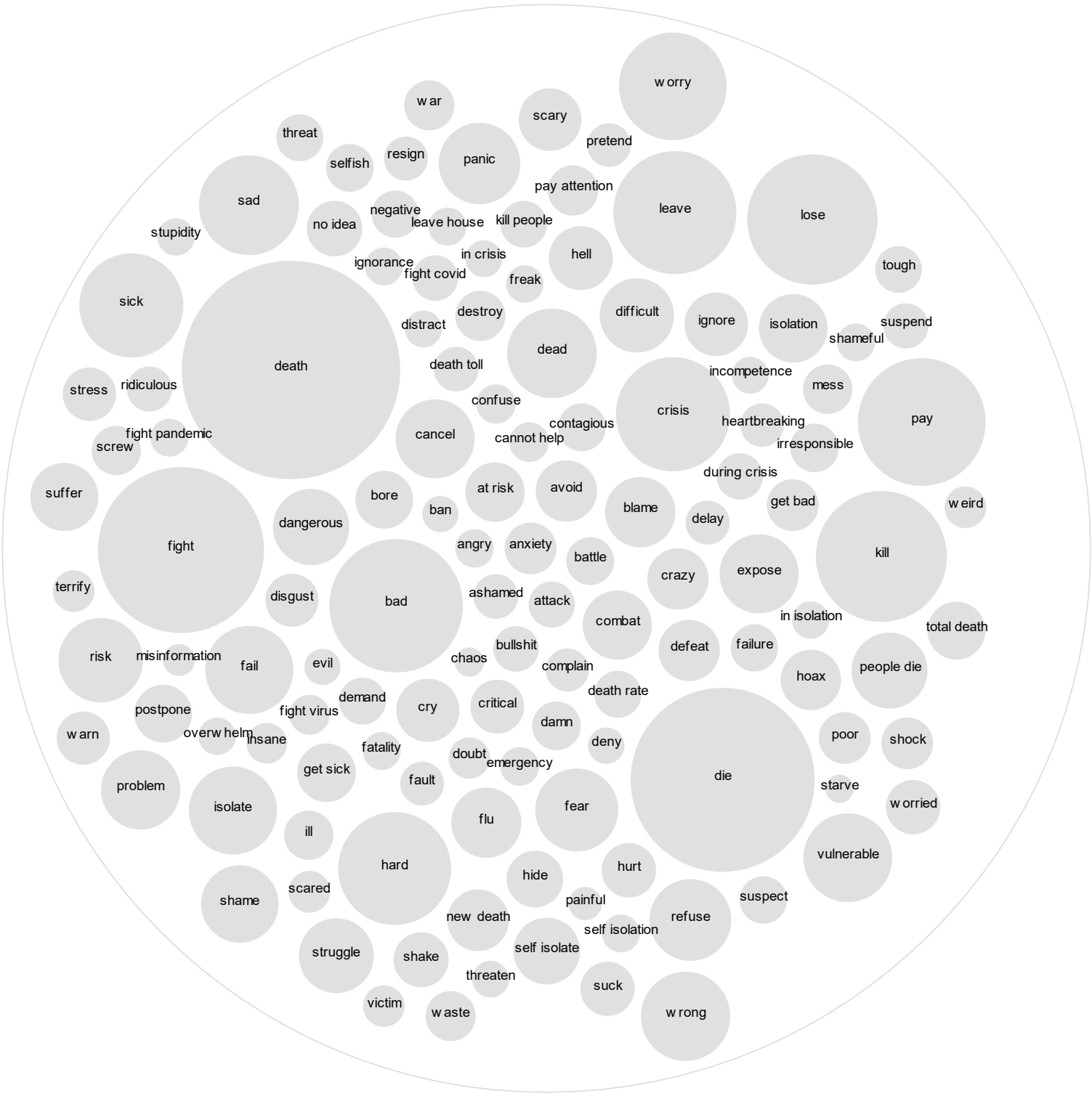

Figure 3. Top 130 negative themes and their dominance in terms of frequency of occurrence (larger size of the gray oval represents more dominance)

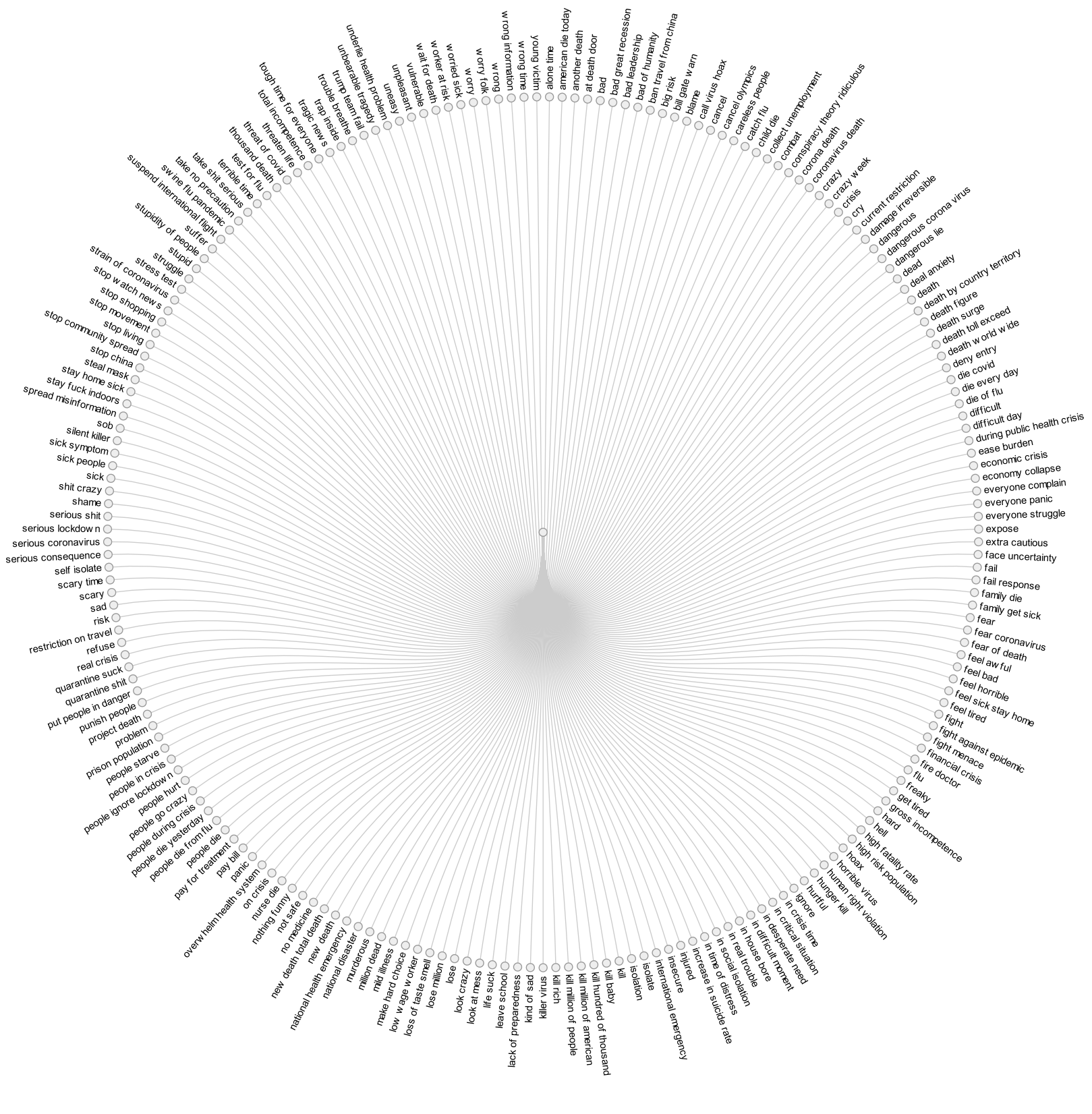

Figure 4. Sample negative themes

## Positive Themes

Figure 5 illustrates top 130 positive themes and their dominance in terms of frequency of occurrence (larger size of the gray oval represents more dominance in the figure). Our results revealed that *help* (n=18,498) is the dominant theme, followed by *hope* (n=7,708), *protect* (n=7,130), *love* (n=6,895), *support* (n=6,198), *good* (n=5,740), *share* (n=5,187), *care* (n=4,917), *stay safe* (n=4,917), and so on. Figure 6 shows more positive themes, such as *keep everyone safe*, *clean environment*, *trust scientific data*, *create cure*, *economic relief*, *encourage business*, *remain strong*, *good mask*, *social distancing best way*, *generous*, *respect human right*, *help prevent further spread*, *pray for health*, *social solidarity*, *support relief effort*, *protect health worker*, *good immune system*, *practice good hand hygiene*, *speak truth*, *expand testing*, *protect vulnerable people*, *free treatment*, *ease anxiety*, etc.

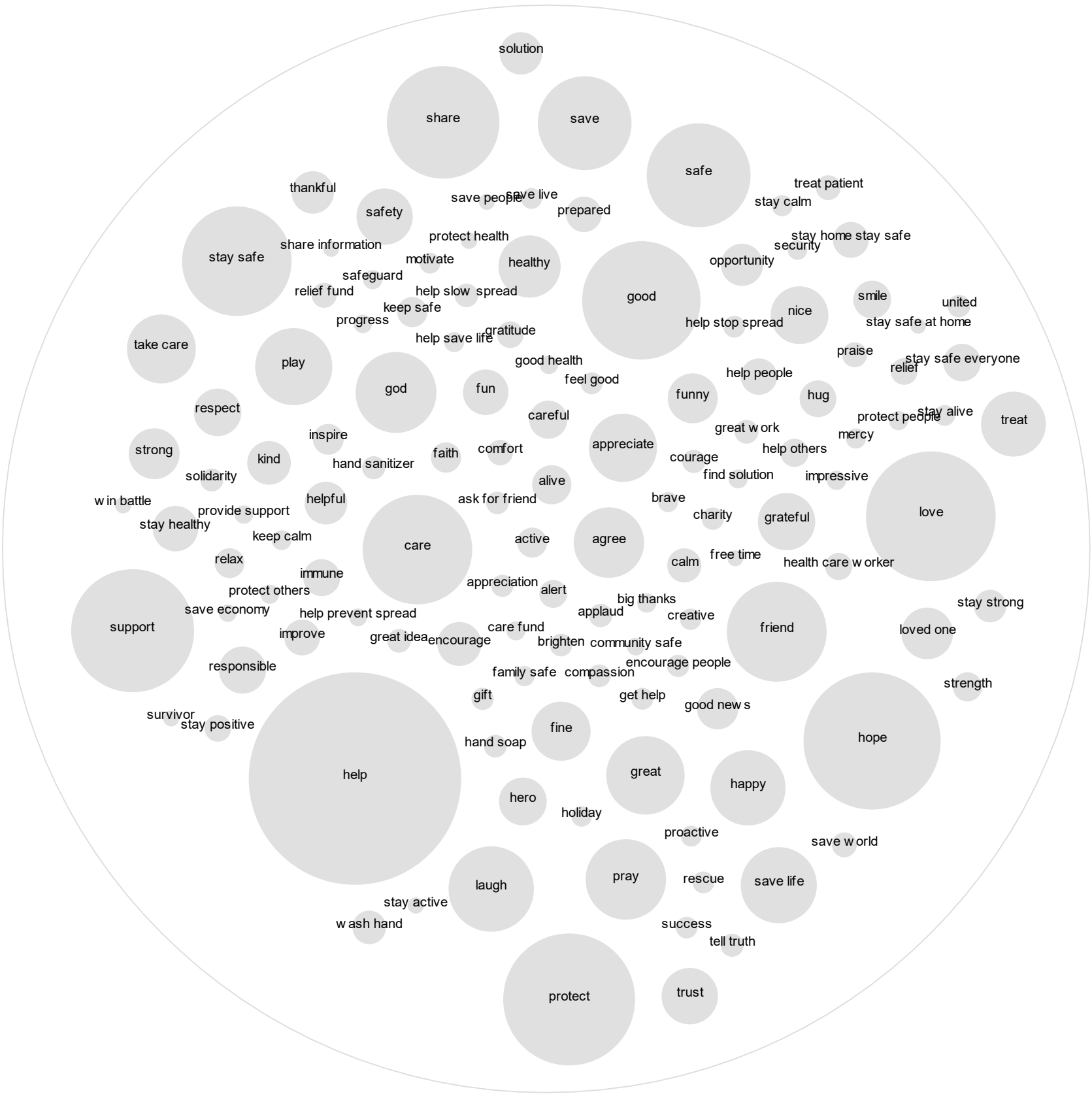

Figure 5. Top 130 positive themes and their dominance in terms of frequency of occurrence (larger size of the gray oval represents more dominance)

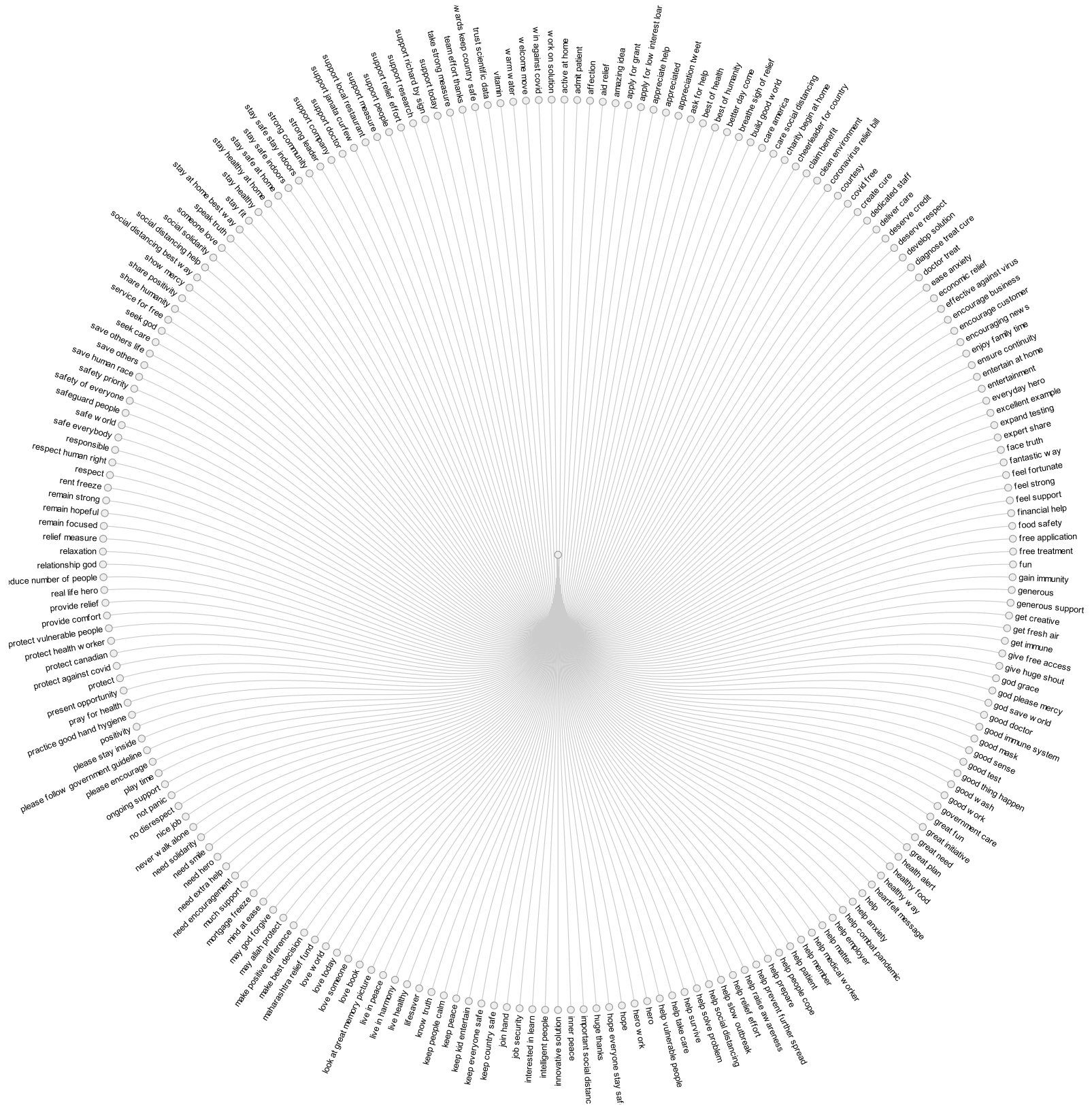

Figure 6. Sample positive themes

## Theme Categories

Overall, 34 negative and 20 positive theme categories emerged after the *Theme Categorization* phase discussed in the Methods section. 15 out of the 34 negative theme categories are health-related, psychosocial, and social issues (which are the main focus of this paper and shown in Table 3). Figure 7 shows the 15 negative theme categories and the corresponding number of themes under each category, while Figure 8 shows the negative theme categories and the total number of comments for each category. *Frustration due to life disruptions* emerged as the top negative theme category with the highest number of comments, followed by *Increased mortality*, *Comparison with other diseases or incidents*, *Nature of the disease*, and *Harassment*. On the other hand, Table 4 shows the 20 positive theme categories, description, and sample comments. Figure 9 shows the corresponding number of themes under each positive theme category, while Figure 10 shows the total number of comments for each theme category. *Public awareness* emerged as the top positive theme category based on the number of comments, followed by *Spiritual support*, *Encouragement*, and *Charity*.

We refer to the theme categories as simply "themes" and the various themes under each category as "subthemes" in the remaining part of this paper.

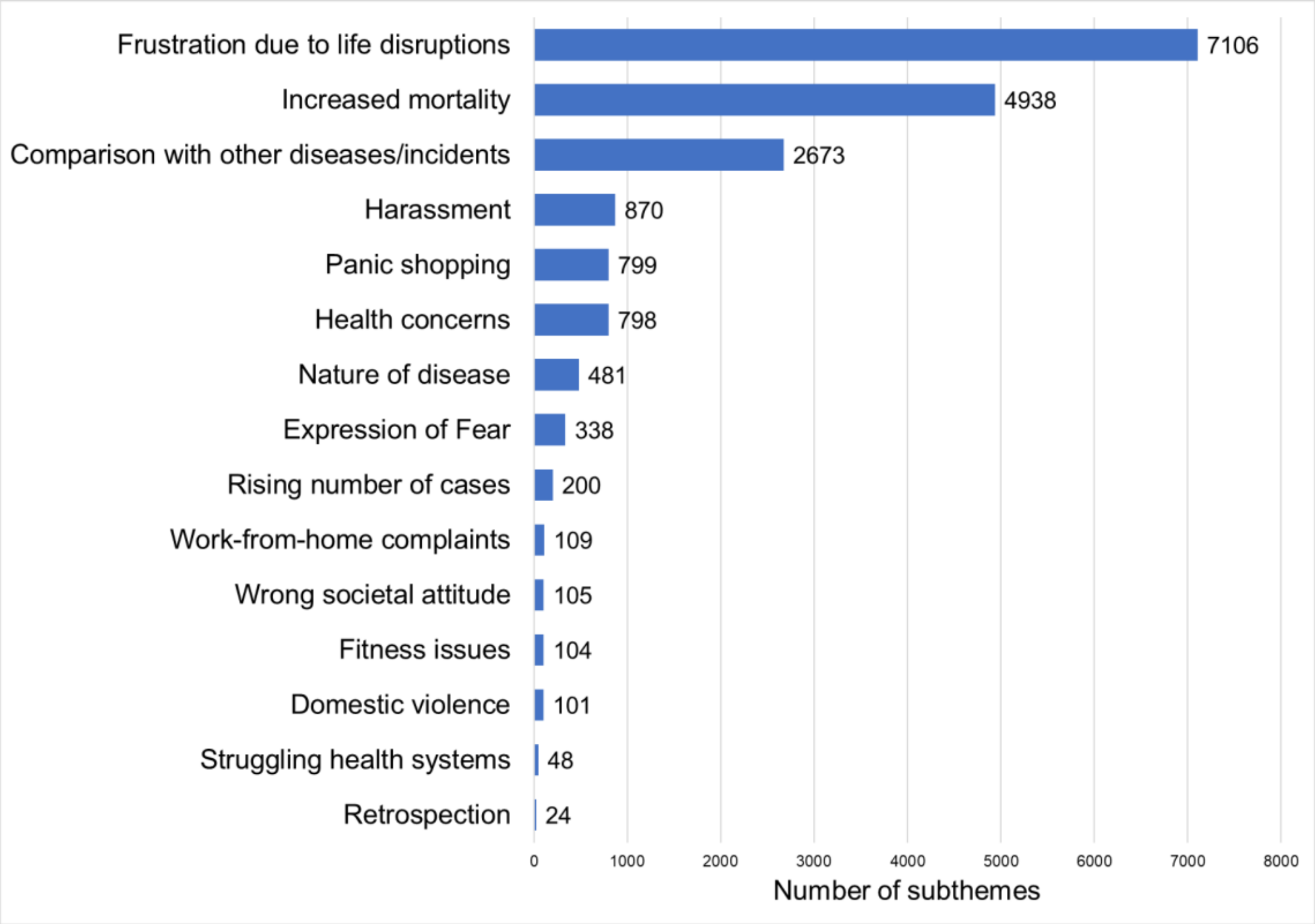

Figure 7. Negative themes and the corresponding number of subthemes

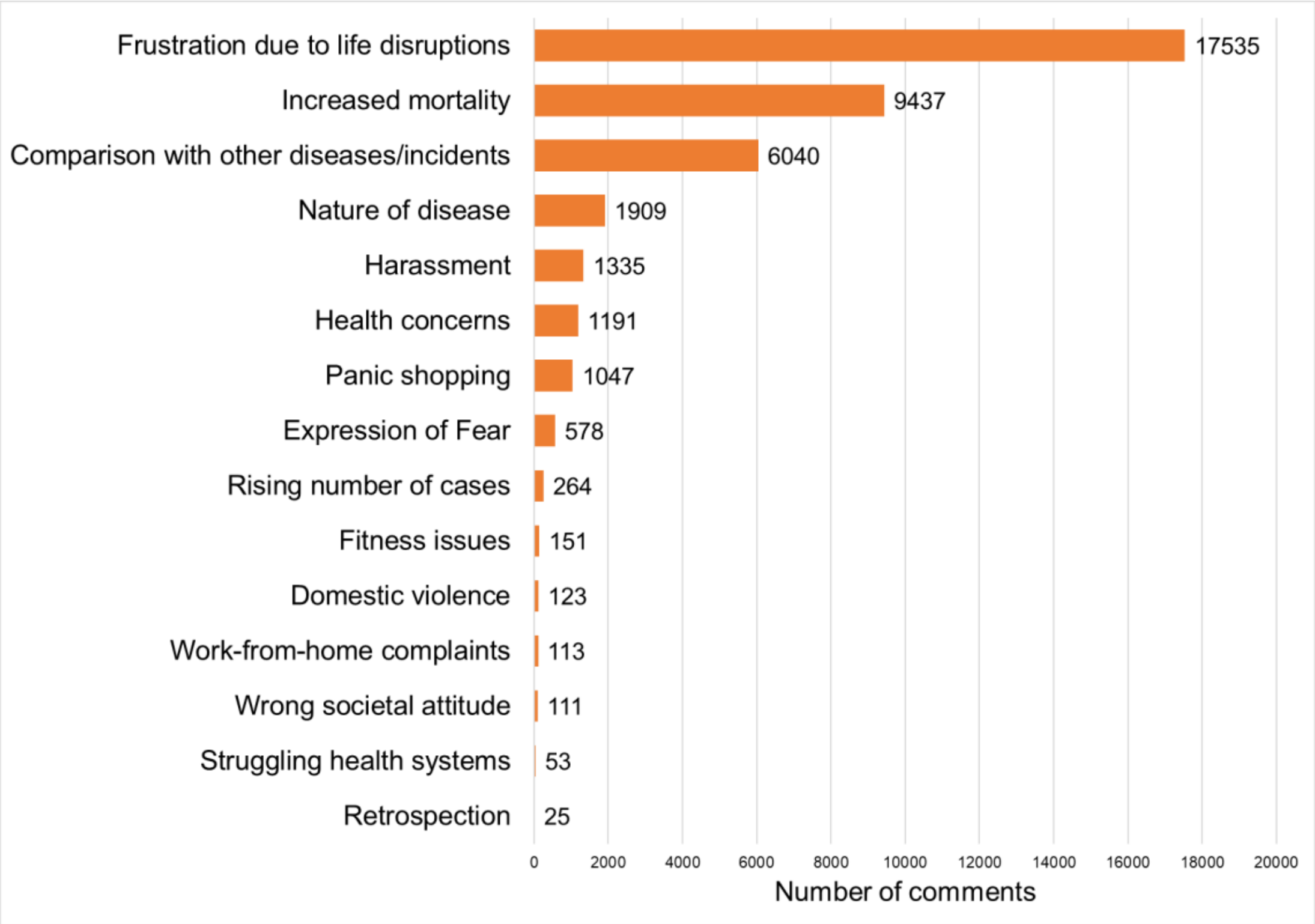

Figure 8. Negative themes and the corresponding number of comments - frequency

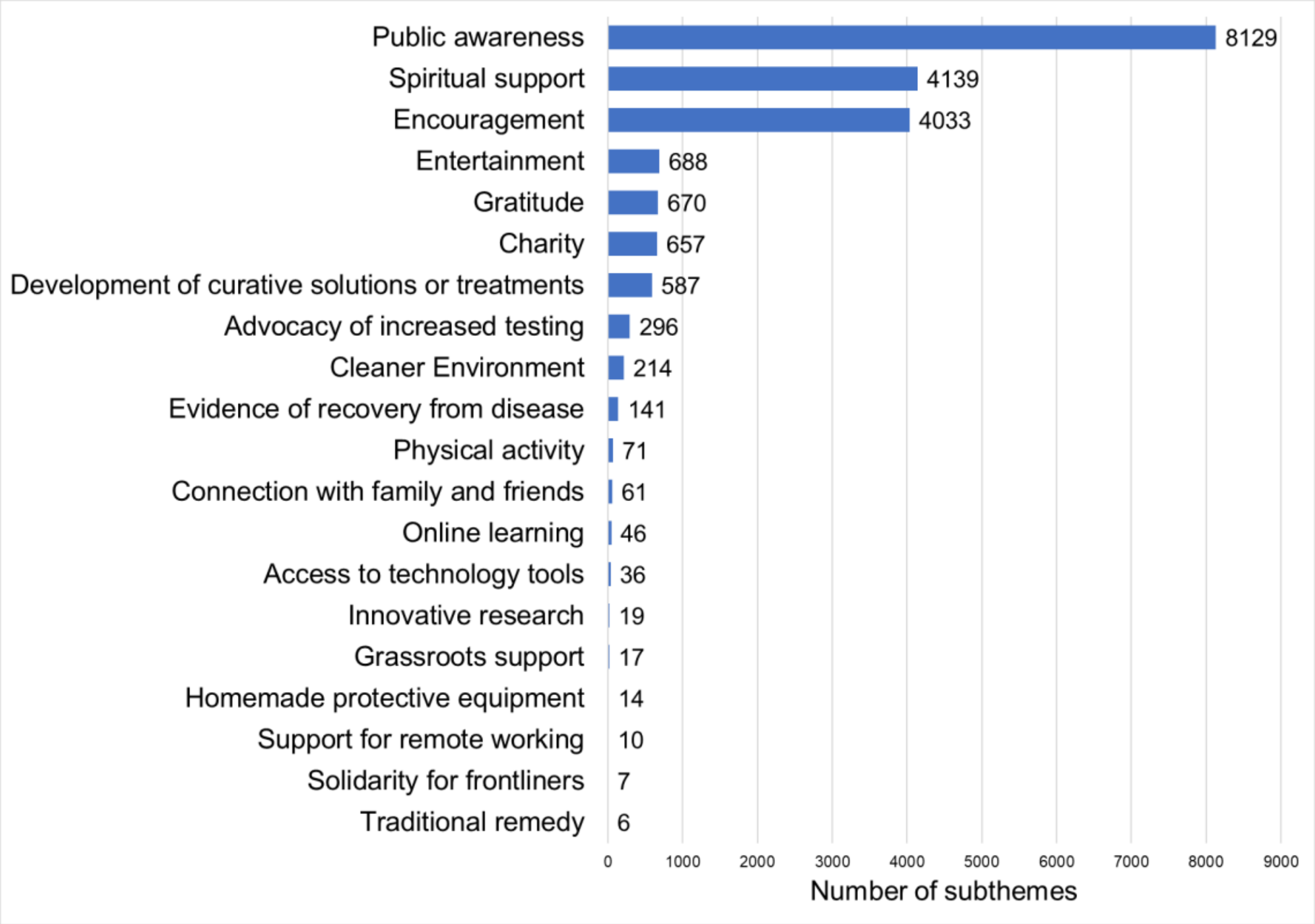

Figure 9. Positive themes and the corresponding number of subthemes

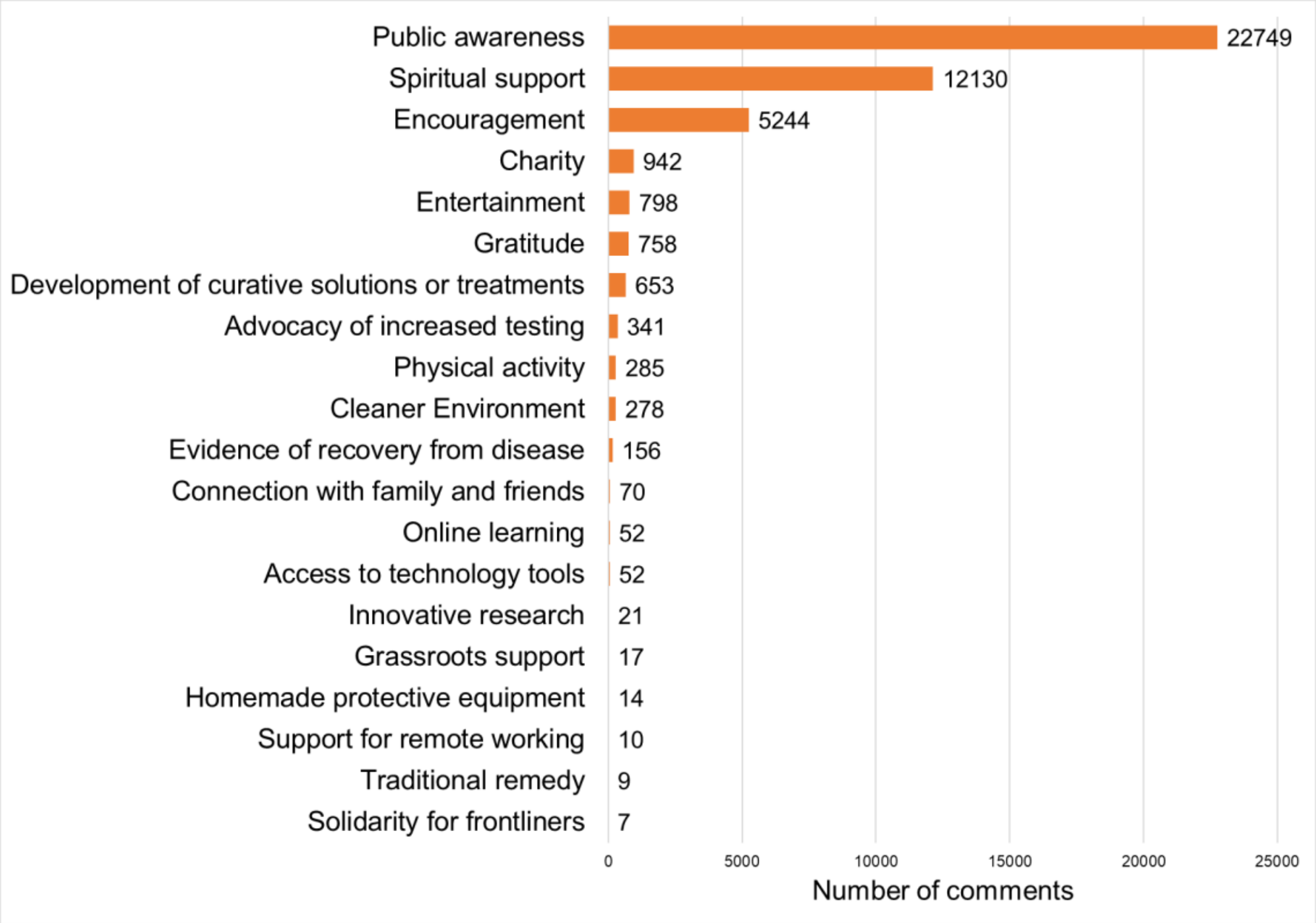

Figure 10. Positive themes and the corresponding number of comments - frequency

Table 3. Negative themes, descriptions, and corresponding sample comments

| | Theme | Description | Sample comments |
|---|---|---|---|
| **Health-related Issues** | | | |
| | Increased mortality | Increasing number of deaths due to the Coronavirus disease | "Grieving for the world and my country, every night the *death count rises up*. I cry for everybody that as died, for those people fighting it, & family who have lost someone. I do not know each person by name BUT I want you to know you are not alone in this pain." [a] [C58]<br><br>"... The number of deaths from the CoronaVirus in London are *doubling every two days*. London could end up with a worse than Italy." [C90] |
| | Health concerns | Health concerns expressed by people, such as mental health issues (e.g., anxiety, depression, stress, obsessive compulsive disorder, etc.), excessive drinking, migraine, fatigue, and others | "It is been either ten or fourteen days since the nursing home I work at went on lockdown due to Covid_19 and the *stress/anxiety is really starting to get to me*. I am *struggling to sleep at night*." [C3327]<br><br>"On my fifth day of sickness, the symptoms disappeared, leaving only an *odd metallic taste* in my mouth, *nasal mucosal ulcers* and *intense fatigue*. This is what a former chair of the UK RCGP went through after catching COVID19." [C945] |
| | Struggling health systems | Inability of health systems to cope with pandemic and give people adequate healthcare | "What clearly shows is the correlation between countries with clean hospitals and countries with bad hospitals and corona deaths. People die with corona not of corona. In particular *New York and California displays its poor health system*." [C81]<br><br>"Pakistani doctors openly saying their numbers are being underreported. They claim 100s of patients in Lahore alone, and say that |
| | | | *because of poor facilities*, hospitals themselves might be spreading the virus." [C44] |
| | Fitness issues | Inability to perform usual physical activity nor attend fitness sessions, and dislike for indoor workout | "Woke up at 8:20 am and still in bed from the past 2 hours. *No mood of workout*" [C271]<br><br>"I just need it to be known that I *hate quarantine workouts* and I miss the damn gym. Also I never thought I would ever say this in life!" [209] |
| | Rising number of cases | Concerns over rising COVID-19 cases | "The United States is currently on the path of the *most widespread viral attack in the world*." [C908]<br><br>"This has done a heinous crime on humanity by *spreading to more and more areas of the country*. The head of the Jamat must be booked for murder crime, as many have died due to Corona infection after attending their function." [C24] |
| | Nature of disease | Explaining the nature of the COVID-19 disease, including its symptoms (such as cough, loss of taste and fever), its "no symptom" (or asymptotic) behaviour, how it spreads, and vulnerable population | "Recent reports suggest that Covid-19 does not only *affect the respiratory system*, but also *affects the Central Nervous System*. *Loss of smell* and *loss of taste* happen to be some of the early symptoms of covid-19." [C660]<br><br>"…they are all *grabbing and touching the desk*. This *virus is much more contagious* than the flu and there is 0 *immunity exposure illness*. Keep doing it like this and it is only a matter of time before everyone who spoke will have the virus…" [C900]<br><br>"If 3 elders died within 72 hours and *had no symptoms*, should we not be testing everyone?" [C2447] |
| | Comparison with other diseases or incidents | Comparing Coronavirus disease with other diseases or incidents, such as flu, pneumonia, natural disasters, war, etc. | "A *war fought with no guns or bombs*, where people flee from what they do not see or is it World War 3?" [C515] |
| | | | "I believe Covid19 is this mutated *H5N1 avian flu virus*. It is airborne, which might explain the rapid spread around the world." [C671] |
| **Psychosocial Issues** | | | |
| | Expression of Fear | Expression and spread of fear among people, including fear of infection, sickness, and death | "… Indigenous tribes are closing off their reserves to visitors as they *fear the disease* that is fast spreading across South America could wipe them out…" [C3884]<br><br>"…This virus has caused *a lot of fear in the lives of many*, it has also brought about different mindset in the heart of men. Truly the world is coming to an end." [C7227] |
| | Retrospection | People recalling past life prior to pandemic and wished they got it back | "I *miss football*. I *miss my family*. I *miss my friends*. But more than all of that, I *miss human touch*!" [C300]<br><br>"If you consider yourself and how is social distancing going for you? I *miss walking around and reading/writing at local coffee shops*." [C3001] |
| | Frustration due to life disruptions | Expression of frustrations over disruption to everyday life, such as consistent homework (for schoolers), more household chores (for moms), difficulty accessing family members or loved ones, sporting event suspension, postponement of planned trips and tours, higher food prices, restaurants closure, etc. | "Everyday, I wake up from a very normal dream, and realize I have to do another day in this insane world. *All I want to do is go see my mom and give her a hug, but I cannot*! *I feel so alone*. I *cry every day*. I cannot do this much longer." [C2905]<br><br>"Day 15 of I only leave my house for food and exercise. *Living in such extremes is confusing and disorienting for my body*. Every time I step outside, I become hungry and start sweating, preemptively." [C4484]<br><br>"*Not getting a hair cut in February was a terrible idea*. I am two hair shades away from looking like an overweight member of BTS" [C89] |
| | Work-from-home complaints | Complaints about working from home during pandemic, such as distractions/disturbance, psychological stress, pain, sleep issues, etc. | "Is anyone else experiencing *leg and knee pain from working from home* too much or is it just me? If so, how have you all dealt with it?..." [C824]<br><br>"*Working from home is an epic fail*! I am losing it between my child, pets, monkey calls, and renovations. Wake me when this is over." [C853] |
| | Panic shopping | People stockpiling groceries and other essential goods due to pandemic | "This *panic buying is ridiculous*! Heart rending! Guess it takes a situation like to show how selfish callously indifferent we really are towards other humans / animals. Have a heart people!" [C779]<br><br>"And yet there is still no logical reason for *clearing the shelves* of toilet roll, depriving those who are old infirm or in poverty from accessing such necessities because the self-serving privileged have greedily taken it all away." [C4] |
| **Social Issues** | | | |
| | Wrong societal attitude | Blaming people's wrong attitude as causing or fanning the spread (such as disobeying instructions from governments and healthcare professionals, eating bats and wild animals, etc.) | "Cannot believe how *irresponsible people are being in regard public health*. We all have a duty of care to each other, please abide by the social distance rules" [C4924]<br><br>"...We are only *going to die from our own arrogance* if people keep going outside gathering, believing they will never get it or infect others..." [C2557] |
| | Domestic violence | Rise in domestic violence cases in homes | "Of all women murdered with a gun in the US, half are killed by their intimate partners. *COVID19 pandemic is causing a rise in domestic violence*. Close the gun stores." [C56]<br><br>"… a police station in China received 162 reports of *domestic violence* in February compared to 47 for the same |
| | | | month in 2019. Advocates attribute this *rise in cases to the lockdown*." [C100] |
| | Harassment | Harassing and blaming people from certain countries, race, or religion as responsible for the Coronavirus disease | "… stop being so *racist against Northeastern state of India*. We are not Corona. Get your damn information right. The most affected places with is not Northeast. Instead of being a racist and criticising others, why do not you be more careful?" [C1413]<br><br>"An Asian American couple in Minnesota found a *racist note taped to their front door blaming them for the coronavirus*" [C921] |

[a] All comments are included verbatim throughout the paper, including spelling and grammatical mistakes.

Table 4. Positive themes, descriptions, and corresponding sample comments

| **Theme Category** | **Description** | **Sample comments** |
|---|---|---|
| Gratitude | Appreciating health workers, delivery workers, farmers, pilots, security agents, and other frontline worker, as well as the government, for their active role during the pandemic | "Let us show our *appreciation for hard work the health care professionals* are doing to save lives at these thought times." [C37]<br><br>"*Thank you to all on the Frontline and the Key Workers*, keep up the amazing work, you are doing an amazing job, keep our country going. Keep Smiling in challenging times" [C156] |
| Public awareness | Raising awareness intimating the public about general safety and control measures to limit the spread of the disease (such as good hygiene, social distancing, staying at home, face masks, healthy eating, etc.), addressing misinformation, travel guidelines, etc. | "Practicing *good hygiene*, like *washing your hands often*, *with soap and water*, *for at least 20 seconds*, is the best way to prevent the spread of Coronavirus..." [C707]<br><br>"We can help ourselves by keep *practicing proper hygiene, assume anyone around us could be positive* and we must *keep personal distancing*, *wear protective gear*, *mask*, *gloves*, etc. Boost our immune system by *eating healthy food*. Search what kind of *vegetable* and *fruits* is best to increase our body's ability to fight the virus. *Sanitize groceries* before bringing them inside the house..." [C16]<br><br>"If extremely necessary, *stay healthy while travelling by maintaining personal hygiene*, *cough etiquette* and *keeping a distance of at least one metre* from others. Here are some travel tips from World Health |
| | | Organization" [C537] |
| Cleaner Environment | Evidence of cleaner environment, including less pollution and good air quality, due to pandemic-related lockdowns | "Coronavirus, *making Earth healthy again*" [C588]<br><br>"Coronavirus pandemic leading to *huge drop in air pollution*" [C664]<br><br>"Remember the time when we used to share the post that said, 'We have only 6 months to take action and save the environment, mend your ways or there is no way to save the earth'. Guess what? *A virus saved the environment* better than the most evolved beings did." [C992] |
| Evidence of recovery from disease | Evidence of people recovering from Coronavirus disease with or without treatment | "A 95-year-old, become the oldest woman in to *recover from the novel coronavirus without the need for antiviral treatment* after her body showed a great reaction to the disease, doctors say." [C214]<br><br>"He was referring to the number of patients treated in the Baptist Healthcare system here. He said the numbers show *90% of their Coronavirus positive patients recover at home*." [C321] |
| Homemade protective equipment | People's ingenuity in creating essential protective equipment (e.g., face masks, face shield, etc.) themselves in their homes or community | "When your missus tells you she put Vodka in her fairy. How was I supposed to know she is *trying to make homemade hand sanitizer* with the washing up liquid?" [C9335]<br><br>"82 and counting. Eager volunteers all over the city are *stitching face masks to ensure that our community remains protected* despite the imminent shortage of PPEs in this pandemic." [C129] |
| Online learning | Public engagement in online learning, such as schools teaching students/pupils online, self-development by enrolling in online courses covering different domains, etc. | "We will all be spending more time at home in the next few months due to coronavirus. I have already *registered myself with number of short courses with online learning* for the subjects that interest me. This could be a good time to learn something new or sharpen up your skills." [C1648]<br><br>"Despite all the news, today is the first day back to *school for Florida's students virtually. Distance learning* might be the new normal for a while..." [C710] |
| Connection with family and friends | Spending time with family, friends, and loved ones due to pandemic and lockdown | "Yesterday my mom had a *video call with her group of friends* instead of going out for their regular meetup. I am proud of my mom. Who said boomers are outdated?" [C7151] |
| | | "This is not the time to be selfish. This is the time to be more present. I did a roll call this morning, *calling all my friends video call, family and those I care about* just to check on them..." [C2560] |
| Entertainment | People accessing entertaining contents online or offline, such as watching movies on streaming websites, playing games, etc. | "Stay at Home and Stay Safe. Please share me good *Amazon Prime or Netflix movies and television shows*." [C93]<br><br>"Downloading the *play station 1* and *nintendo 64* emulators on my pc. Can get any game." [C399]<br><br>"9th grade Colin has returned and will be *playing grand theft auto* for 8 hours straight" [C3192] |
| Charity | Provision of relief packages (including donations, gifts, and fundraising) for individuals, businesses, and hospitals to ease financial burden caused by the pandemic | "Kicking off NeighborsHelpingNeighbors here in RI with the *RhodeIsland Hospitality Relief Fund* - a fund aiming to help industry members directly affected by COVID19." [C3623]<br><br>"The 2 trillion *emergency relief package* now before the House provides the following: 1,200 *checks* for those earning less than 75k, plus 500 per child, *unemployment benefits* for 39 weeks up from 26, unemployment benefits rise by 600/week for 4 months" [C10913]<br><br>"Zlatan Ibrahimovic has *created a fundraiser* to gather 1m to help hospitals in Italy working to tackle the coronavirus. The striker has *donated €100,000* to get the fund started." [C24] |
| Advocacy of increased testing | Advocacy of increased testing as a means of curbing spread by detecting infected people and isolating them quickly | "Studies in Iceland show that half of carriers show no symptoms. Having *widespread test allows them to isolate those with the virus*, and thus, the virus itself. *Test, test, test*." [C1333]<br><br>"US surpassing all countries in number of patients. Is it simply because they are now *testing at a faster rate* even on slightest suspicion? Yes, one of the ways to stop this pandemic is *Testing, Testing and Testing* as recommended by WHO" [C6363] |
| Grassroots support | Extending support to people at the local or community level during the pandemic | "The amazing QueerCare are offering *support to local mutual aid groups*. They have trained volunteers and are already in *contact with people who will need support over the weeks and months to come*..." [C2839] |
| | | "My self, and some other good citizens residing in Lagos *bought 150 pieces of pocket hand Sanitizers to distribute to people who cannot afford it* or have knowledge of what is all about and in same process to tell them how to stay safe. That is our own giveaway." [C773] |
| Access to necessary tools | Access to tracking or communication tools/features for information dissemination or remote communication during pandemic and lockdown | "This is the most *stunning visualization* of how spread around the world. A mesmerizing and terrifying display of globalization and virus spread." [C6625]<br><br>"I would like to personally thank for the *private chat function* during Zoom video meetings." [C2244]<br><br>"This *tool* is useful in identifying those who are medically more at risk of suffering complications from COVID19 ..." [C9233] |
| Spiritual support | Offering prayer of recovery for those with pandemic-related health conditions and those at risk, as well as a show of hope in challenging times | "I *pray for everyone diagnose of COVID19 swift and complete recovery* in Jesus name. Dear Lord, heal the world. Take away and give everyone good health." [C9230]<br><br>"Let's *pray for all African communities* who live in hostels, huts, rural areas with no connectivity. No information. Innocents but will also be affected by the CoronaVirus" [C2104]<br><br>"...those who have relationship with God are less likely to become depressed than those who do not. It is because their *confidence and hope is in Him* so, *let us trust God amid the pandemic*." [C1349] |
| Solidarity for frontliners | Public call for support and protection of frontliners, such as health workers, delivery workers, etc. | "What is New York City doing to *protect the workers at Amazon fulfillment center*? The virus is spreading quickly among the employees…" [C3332]<br><br>"While the news of is getting more urgent by the hour, it is great to know that during a turbulent time some corporations showed their support by *increasing the minute wage for frontline workers*, *hiring 100s of people* to assist with increased demand, helping unite." [C4439] |
| Development of curative solutions or treatments | Ongoing efforts by health researchers to develop vaccines or drugs to cure or treat the Coronavirus disease | "There is *a bunch of solutions* being researched from I guess *infusions to drugs to slow virus reproduction*, to *vaccines*. None will be ready until a year or two. Clinical studies take time and that is how we do medical science safely." [C5759] |
| | | "Coronavirus *vaccine clinical trial starts* Monday, U.S. official says…But officials still say it will take a year to 18 months to fully validate any potential vaccine." [C2034] |
| Physical activity | Efforts made by people to stay active and fit, as well as physical activity suggestions, during lockdown or isolation | "I know it is a bad situation but damn I am *improving my fitness* during this lockdown" [C6335]<br><br>"Just finished our *fitness session* via skype with friends. I live in Barcelona, lost track of how many days since I went outside, we have to *keep the body moving* though" [C9430]<br><br>"*Get active every day* with the kids or by yourself! GoNoodle is here to help!..." [C9364] |
| Encouragement | (1) Encouraging people to stay calm as they cope with the pandemic situation, (2) Encouraging people to view the pandemic from a positive standpoint and stay productive amidst the challenges, (3) Encouraging people to help others in need and not panic buy, and (4) Encouraging people to obey lockdown rules, as well as guidelines released by governments and health professionals | "Let us all *stay calm*. Give the *authorities time to attend to and address public concerns*…" [C3]<br><br>"Yes, we can. If we *stay calm* and *respect the rules*, together we will defeat the enemy." [C38]<br><br>"*Stay calm* and *help each other*. Be *careful*. Do *not panic buy*, and *never give up*!" [C164]<br><br>"We have done it before and can do it now. See the *positive possibilities*. Redirect the substantial energy of our frustration and turn it into *positive, effective, unstoppable determination* for our *safe and healthy future*" [C2273] |
| Support for remote working | People's support for the work-from-home measure, including adapting/coping with the challenges it brings | "*Working from home* with the children of school can be challenging. We have taken a look at some of the *ways you can help structure your day and stay on top of working from home* with our schools closed." [C148]<br><br>"*Working from home amid Coronavirus pandemic was amazing* at least today. Came up with some amazing designs…We are coming up with our first property in June. Already sold out" [C53] |
| Innovative research | Global research efforts to create innovative products to address the pandemic, including developing interventions (such as digital or technological interventions) that help people socially, physically, emotionally or psychologically, as well as | "Doctors and scientists...have *designed an application* to help the public *monitor their symptoms and the spread of the virus in real-time* with the contributing to their own vital research" [C96]<br><br>"...Well, it is a huge scientific discovery! Scientists want to use *artificial intelligence* |
| | improve their overall health and wellness | *technology* for a quicker and cheaper COVID-19 screening...” [C197]<br><br>“The COVID19 Global Hackathon is an opportunity for *developers to build software solutions that drive social impact*, with the aim of tackling some of the challenges related to the current coronavirus pandemic...” [C619] |
| Traditional remedy | Some suggestions regarding the natural or traditional means of protecting the body from contracting the disease | “*Gargling vitamin c, vinegar, warm water*, and a *little bit of baking soda every 20 minutes*. After 5 days, she tested negative. If you or anyone you know starts getting symptoms, this can help! Catch it early before it gets to your lungs!” [C803]<br><br>“...I am sure I have Covid_19. I believe the *natural healing helped my daughter but suppressed my symptoms*...” [C2777] |

## Discussion

### Principal Results

In this paper, we analyzed social media comments to uncover insights regarding people’s opinions and perceptions toward the COVID-19 pandemic using context-aware NLP approach. Our empirical findings revealed negative and positive themes (see Tables 3 and 4) representing negative and positive impact of COVID-19 pandemic and coping mechanisms on the world population. We discussed the implications of the negative themes in this section and then recommend interventions that can help tackle these issues based on the positive themes and other remedial ideas rooted in research.

### *Negative issues surrounding COVID-19 Pandemic*

Table 3 shows the negative themes grouped under health-related issues, psychosocial issues, and social issues from our results. The health-related issues include *health concerns, increased mortality, struggling health systems, fitness issues, nature of disease, rising number of cases,* and *comparison with other diseases or incidents*. Psychosocial issues are *expression of fear, panic shopping, retrospection, work-from-home issues,* and *frustration due to life disruptions*. Social issues are *wrong societal attitude, domestic violence,* and *harassment*.

#### Health-related issues

Evidence shows a rapid increase in the number of COVID-19 cases and a high case-fatality rate of 7.2% [68]. In addition, substantial number of infected patients had severe pneumonia or were critically ill [68]. Another evidence revealed the mental health issues experienced by people and health professionals directly impacted by COVID-19 pandemic [69], as well as the global healthcare systems’ inability to deal with the outbreak [70]. The themes under this category are discussed in the following subsections. They align with existing research and uncovered additional insights with respect to the health-related issues caused by COVID-19 and witnessed by people globally.

### Health concerns

Based on our findings, people experienced various mental health issues (such as anxiety, depression, stress, obsessive compulsive disorder, etc.) during the pandemic. This is possibly due to the length of time spent staying at home (which may be traumatic for some people while causing loneliness for others), worrying about being infected with the disease and difficult living conditions, as well as guilt on the part of healthcare workers who feel responsible for being unable to save their patients from death. Research confirms that worry is associated with anxiety and depression [71]. Cases of mental health disorders linked to COVID-19 have also been reported [72]. Furthermore, people expressed other concerns like excessive drinking, migraine, chest pain, mild to severe fatigue, nasal mucosal ulcers, sleep disorder, and others. Sample comments are shown below:

> "*Cannot sleep. Mind is racing*. Feeling *anxious*." [C6648]

> "I am *so stressed* and *my anxiety has hit the roof*. I am anxious about money and how we will cope?" [C238]

> "This coronavirus outbreak is *more stressful for the family*. Doing my best to keep sanitized and safe. But, fear of the invisible killer lingers on, *taking a mental toll on my mother, wife, son*, who are petrified every time I walk out of the main door." [C116]

> "The *chest pains* today is beyond. It kinda have crawled up a bit and I feel like I put my hand on my heart from time to time. *Very tired* today. But weirdly still no fever. But I am *cold* and I *feel sick*." [C12293]

> "We have only been in for 3 weeks we are already *feeling anxiety*, *depression severe,* so we decided to think of some ways we can keep ourselves each other in good spirits." [C8263]

### Increased Mortality

People attested to an increase in death rates in many countries across different continents including North America, Europe, Asia, Middle East, and Australia, as shown in the sample comments below. Many countries, especially those in Africa, started reporting deaths from COVID-19 (see [C1264]). Our findings also revealed people of varying demographics died from the disease, including teenagers, adults, and older adults, as well as those with or without underlying health conditions (see [C8837] and [C940]).

> "This is why *America leads the world in the death toll already*, and the pace still is not showing any signs of slackening." [C3399]

> "*UK coronavirus death rate DOUBLES as 381 die in 24 hours* and *boy, 13, with no health problems becomes youngest victim*" [C8837]

> "Turkish health minister shares latest coronavirus data: *16 more people have died*, bringing death toll to 108" [C9000]

> "*Kenya has recorded the first death*. According to Health Cabinet Secretary,...the 66 years old man died on 26th in the afternoon at the AghaKhan hospital intensive care unit." [C1264]

> "100 more UK deaths in last 24h alone. These are *not all elderly co-morbid people*. Among these are the *young*, and *the fit*..." [C940]

### Struggling health systems

Health systems around the world are struggling to cope with the surge in the number of COVID-19 patients, and in most cases are unable to admit patients due to limited resources [73]. Research has shown that healthcare burden due to COVID-19 is associated with the increase in mortality rate [74]. As revealed in the sample comments below, our findings corroborated evidence of overstretched global health systems during this pandemic.

> "Hospitals turning away coronavirus patients in California. EMTALA being used to set up tents outside of hospital ER, then dumping patients without treatment or testing. I *obtained recordings of standard hospital practice of rejecting Covid19 patients*." [C4949]

> "These are not just numbers. These are people and families. These lives can be saved if the chunk of $2.2 Trillion are not used for bailing out corporations and used to fix the *broken US Health System*." [C3000]

### Fitness issues

Evidence argues the prevalence of physical inactivity globally due to nationwide quarantine or lockdown [75]. This is confirmed in our findings which show that people have trouble staying fit due to inability to control eating habits or urges while at home, as well as personal dislike for indoor-only workouts, as shown in the comments below. Physical inactivity has been linked to coronary heart disease, diabetes, stroke, and mental health issues [76–78] which, in turn, are risk factors for mortality in COVID-19 adult inpatients [79].

> "…severely missing my gym, missing routine, and *cannot control my eating while at home*. Things are getting bad." [C9002]

> "During this shelter in place, I was gonna eat healthy and kill some workouts. *But instead I've been Guy Fieri'ing around the kitchen sampling all my quarantine food every 2 hours*. At this point, which will get me first - coronavirus or a coronary?"

### Nature of disease

People expressed their opinion about the nature of COVID-19 based on their experiences and information available to them. As shown in the sample comments below, people with underlying health conditions (e.g., diabetes, heart disease) are at higher risk of developing severe complications from the disease. In addition, the asymptotic attribute of COVID-19 is also discussed, and the possibility of the virus to infect some critical immune cells that may lead to the failure of sensitive organs like the lungs. People also perceived the disease as racial- or nationality-independent but seems to pose more risk to men than women. The disease is also seen as highly contagious and shows symptoms such as cough, fever, fatigue, loss of smell, muscle aches, and respiratory-related symptoms (e.g., shortness of breath). These findings align with clinical evidence regarding COVID-19 [80–85].

> "Some WTC Health Program members with *certain health conditions*...may be at *a higher risk of serious illness* from COVID19" [C3620]

> "…Majority *do not show symptoms* while spreading Covid_19" [C10033]

> "...If the coronavirus *infects some of the immune cells* then there is a cellular catastrophe and *organs like lungs are gone within hours*." [C9200]

> "Coronavirus is a disease that *pays no attention to borders, race or nationality*. However, it appears COVID-19 does *pose a noticeably bigger threat to men than it does to women*." [C1902]

> "This is absolutely true. If you have a combination of *cough, fever, problems smelling, weakness, muscle aches or shortness of breath*, assume you have covid19. Don't bother getting tested. I know of many docs who don't test anymore if patient has obvious symptoms." [C2729]

### Rising number of cases

Our findings show that more people are getting infected with COVID-19 in many parts of the world, as shown in the sample comments below. Evidence confirms increasing number of COVID-19 cases in North America [86,87], Europe [88], as well as a growing concern for vulnerable continents such as Africa [89].

> "There is *rapid increase in cases of COVID19 in India*...I request PM to extend the lockdown to avoid community spread." [C1325]

> "Despite *infection cases increasing at exponential rate* doubling every 3 days, Trump pushes workers to risk their lives for economy..." [C6522]

### Comparison with other diseases or incidents

Our findings reveal that people compare COVID-19 with other diseases such as flu (e.g., Spanish flu and H1N1 swine flu) and SARS, as well as with more extreme incidents such as war. However, while some people tend to downplay the severity of COVID-19 (see [C647]), others think it is dangerous or frightening (see [C922] and [C45]). Research shows that COVID-19 has a higher transmissibility rate than SARS [90] and killed more people than SARS and MERS combined [91], thereby making it a highly contagious and lethal disease.

> "It is just *another strain of flu*. People with weak and have health problem it will affect different than people with stronger and not so much health problems. *Media is making it sound worse*..." [C647]

> "I was not at all concerned about swine flu, I was not at all concerned about Ebola, I was not at all concerned about Zika virus, but this virus *I was concerned about going all the way back to January*. If I had enough information to be concerned about this virus, then so did they. We are a month behind on dealing with this virus and there are no excuses, even a lay person like myself knew all the way back in January that this was *bigger than anything that has come before it in my lifetime*..." [C922]

> "*This is a war*. We need to protect ourselves and minimize unnecessary contact to avoid another Spanish flu that killed 50 million people..." [C45]

## Psychosocial issues

### Expression of Fear and Panic shopping

Based on our findings, people are fearful or scared about COVID-19 and while many expressed genuine fear (including those who had lost loved ones to the disease, contracted the disease, or had an infected family member), others attributed it to fear mongering that is further amplified by the media. As a result of this fear, many people engage in panic buying to stockpile essential items so they can stay indoors/limit movements for some days or weeks to keep themselves and their families safe. Below are sample comments:

> "*Very frightening* when people who have travelled think that covid cannot affect them. Such foolish behaviour and thoughts *putting all of us at risk*. Those who travelled please STOP moving around and be at home." [C8887]

> "*Fear mongering* through projecting number of possible deaths. The media is disgusting." [C5559]

> "Everyone who is *panic shopping* is driving my family and me nuts...Everyone in our area is *panic-buying groceries*. We *can't get noodles, rice, or really any real staple foods*. We hardly have any food already. I'm kinda scared." [C2937]

### Work-from-home complaints

Furthermore, the pandemic triggered work-from-home (or remote work) measure to promote continuity of businesses during lockdown [92], but this may have negative implications on people's lifestyle and wellbeing. For example, people found consistently working from home exhausting, boring, and distracting with kids at home. In addition, people living in countries without stable electricity and strong internet found it difficult and more costly to work from home, as they have to fuel their generators and pay more for considerably good internet connectivity. Evidence shows that people work longer hours at home than onsite due to difficulty in maintaining clear delineation between work and non-work domains [93], thereby leading to work-family conflict and strain [94]. Below are sample comments:

> "Do you *find working from home exhausting*? You are not alone. Why is that and how can you combat it?" [C11224]

> "This thing of *working from home annoys*. Waking up at 4 am to have things done before the Boy wakes up. Otherwise you will spend better part of the day *watching cartoons with no work done*" [C9902]

> "It is very clear from today's program how the government is not organized or coordinated with this issue. How should one *work from home with constant power cuts and bad internet?*" [C8855]

### Frustration due to life disruptions and Retrospection

Finally, people are generally frustrated about life disruptions caused by COVID-19 (which is the top issue based on our empirical findings as shown in Figure 8). Based on our findings, this frustration is mostly due to decreased leisure and interaction with friends/family, authorities' actions/inactions, as well as uncertainty of upcoming situation which leads to cognitive dissonance [95], insecurity, and mental discomfort [96]. People expressed their frustrations using words reflecting anger and unhappiness/sadness, as shown in the sample comments below. Research shows that positive emotions (e.g., happiness) and life satisfaction decreased during COVID-19 pandemic [97]. Therefore, it is unsurprising that people missed (and crave for) their pre-pandemic lives, in retrospection (see [C377]).

> "I am *getting more and more angry with this current situation we are in*. My favourite show has had to postpone its final episode, after already postponing series 11..." [C7776]

> "We are literally living in a time when arbitrary shit is more important than health, wellness and preservation of life. Entitlement. Ignorance. Selfishness. An untouchable mentality. Humanity at its absolute worst. *April's going to be a painful month to live through*." [C1228]

> “I cannot wait until this whole thing is OVER. *I miss doing nails*. I *miss being in my element and doing the creative things I enjoy*. I have no practice hand; I have no work; I feel lost. I was just licensed in Jan!” [C377]

## Social issues

### Wrong societal attitude

Our findings revealed disapproval and concerns about people’s defiance of precautionary measures or guidelines (such as social distancing and travel guidelines) to curb the spread of COVID-19 (see [C7218] and [C1444] below), as well as some people’s habit of eating animals assumed to be carriers of viruses (see [C4013]). Research highlights certain factors responsible for reduced compliance with public health guidelines, such as poverty, economic dislocation, lack of compensation, and mistrust of science [98–100].

> “The only good thing about Coronavirus is that it will cull the *stupid people from amongst us* - those that *do not take it seriously and continue to gather in public*, those that *go overseas to attend weddings and other events when they know the risk*...” [C7218]

> “The public response to this crisis in the UK has been *absolutely pathetic*. Showing we are an entitled society who cannot handle being told what to do. Embarrassing that *people cannot follow simple instruction*.” [C1444]

> “Why is that someone who knows how *dangerous eating these animals* are and *still go right ahead and eat*. Is it that they do not have any sense or is it just irresponsible stupid people to do this then what are they crazy or just insane?” [C4013]

### Domestic Violence

Furthermore, increase in domestic violence incidents was reported as a result of the COVID-19-related lockdown, as shown in the comments below. Evidence already confirms the link between COVID-19 and the rising cases of domestic violence globally [101–105].

> “*Cases of domestic violence in the USA has skyrocketed* since the CoronaVirus forced couples to stay home together for 14 days or more.” [C9900]

> “While *domestic violence across France increased by 32%* in one week, in Paris it *rose by as much as 36%*.” [C13720]

> “What worries me more than the coronavirus, is the safety and welfare of those *stuck at home with their abusers*, the *children witnessing domestic violence* and the lonely relying on company. Staying home is not always safer.” [C7910]

### Harassment

Our findings also uncovered undue harassment of people from certain culture, race, or religious background while accusing them of spreading COVID-19 disease. Sample comments below reveal public intimidation and racist attack towards Chinese/Asians and certain Indian tribes. This aligns with evidence of widespread anti-Chinese and anti-Asian xenophobic or racist attacks, especially in the United States, both physically and on social media [106–111].

> "...This supermarket *refuses to sell the food to the CHINESE*! We should stop going to this supermarket! We strongly against RACISM towards Chinese people abroad! THIS MUST STOP!" [C2008]
>
> "Coronavirus has only taught me one thing; some *people are so racist*, especially on this platform. The amount of *hate, racist comments* and *abuse* I am seeing *Chinese/Asian people get is painful*." [C5000]
>
> "Indians got racism in their blood, breath, and beyond. A *Mizo girl faced racism* in Pune as a woman kept covering a face whenever the Mizo girl passed by. A friend Naga from maternal side in Mumbai, *got called corona virus* in the middle of an empty road." [C6660]

## *Recommended interventions for addressing the negative issues*

As lockdown and physical distancing persists, people with health concerns should be able to receive medical attention without visiting a hospital. Considering the proliferation of smartphones and the current wave of global digitization, digital interventions using mobile, artificial intelligence (AI), internet of things (IoT), and virtual reality technologies have been shown to be effective for delivering remote healthcare (or telehealth) to patients [112–117]. This is based on our findings under the *innovative research* positive theme (see Table 4) which revealed global research efforts to create digital interventions using emerging technologies to address the health crisis caused by COVID-19. For example, mobile apps that detect mental health issues (e.g., depression and anxiety) based on phone sensors (or wearable sensors) data and self-reports using machine learning/deep learning models, and then guide users through therapeutic procedures or treatments will be useful tools during and after the pandemic. In addition, these apps should allow users to book appointment with doctors/clinicians/therapists and access remote medical advice, diagnosis, and treatments when necessary.

Also, data-driven surveillance systems based on artificial intelligence that predict location of next COVID-19 outbreak can enhance the effectiveness of containment efforts, thereby slowing the spread of the disease and reducing case-fatality rate. Furthermore, the *development of curative solutions or treatments* (see Table 4), can be accelerated by leveraging machine learning/deep learning algorithms. For example, deep learning models can be used to predict chemical compounds that can halt viral replication, as well as to suggest drugs that can be effective against the virus.

To address fitness issues during lockdown, *physical activity* (which is one of the positive themes in our results) programs or sessions with personalized feedback delivered through mobile apps will be helpful. Research has shown that smartphone-based health programs yield significant weight loss and increase physical activity [118]. There is also an urgent need to strengthen the global healthcare systems to cope with current and future pandemics through public and private investments in the health sector on an ongoing basis, such as provision of public health infrastructure that is robust and adequate for the target population and easily accessible, as well as the provision of health insurance for everyone irrespective of financial status.

*Public awareness* (which emerged as the top positive theme in our findings) is also crucial for addressing negative issues arising from COVID-19 by providing timely and accurate information to people, which can be lifesaving. To reach wider audience in an efficient manner and with less cost, public awareness can be delivered through mobile technologies such as mobile-driven and voice-enabled conversational AI agents (or chatbots) with access to evidence-based and clinically validated resources (e.g., precautionary or safety measures approved by public health agencies and organizations, as well as government-approved policies or guidelines) can deliver accurate information regarding COVID-19 to people in their own native language (and in an interactive fashion) through their smartphones. These chatbots can also be made to route difficult questions to health experts for real-time feedback within the same chat session. This will help to improve people's understanding of the disease, including how it differs from other infectious diseases, and how to protect themselves and their families from getting infected with COVID-19. In addition, people will be empowered with information required to effectively respond to fear mongering, domestic violence, and harassment. Evidence already shows the deployment of multilingual chatbot for public health awareness on coronavirus symptoms, diagnosis, and precautionary measures [119]. Furthermore, chatbots can also respond to emergencies by contacting appropriate security agencies and emergency response teams on behalf of the users. Moreover, chatbots can deliver evidence-based therapeutic interventions to people, while coordinating with specialists behind the scenes where necessary.

For people with non-smartphone devices, public health agencies can partner with telecommunication companies to deliver COVID-19-related information directly to their phones as text messages at regular intervals using the short messaging service (SMS). Social media is another platform through which evidence-based information can be shared with the public but may be overshadowed by fake news or false information which is mostly shared on social media [120]. Nevertheless, official COVID-19-related channels managed by (or in conjunction with) reputable international health organizations (e.g., World Health Organization) or local health authorities within the social media platforms, many of which have already been deployed, provide accurate information/updates about COVID-19 cases, fatality rates, and safety measures/guidelines [121,122]. In addition, people receive location-based updates on these channels, including emergency alerts, in a timely and effective manner.

Finally, based on our findings (see Table 4), *connectedness with family and friends*, *encouragement*, *spiritual support*, and *charity* can help to ease people's frustrations, anxiety, and trauma (due to life disruptions caused by the pandemic) by addressing their emotional, physical, and spiritual needs. Evidence shows that psychological first aid and spiritual care can promote a sense of safety, calmness, self- and collective-efficacy, connectedness, and hope, as well as help people confront and overcome fear [123]. Therefore, people should endeavour to frequently communicate and follow up with loved ones (through direct voice/video calls or using social media), encourage others in distress to stay calm and remain positive, identify people's immediate needs and offer necessary assistance, help people find hope and meaning, and ensure safety and comfort of vulnerable population.

Mobile technology can play a key role in facilitating easy access to relief packages. For instance, mobile apps can be deployed with geolocation and multilingual features to help people locate the nearest food bank and charity organizations offering assistance in their geographical area. In addition, charity organizations can effectively mobilize and deliver relief items to more people, including individuals that are indisposed, based on data collected through these apps. Also, the elderly, the sick, and those in self-isolation can indicate their condition while requesting for relief so that their items can be delivered to their doorstep instead of picking it up. These apps can further integrate with other local and international charity organizations to widen the coverage of relief efforts. Recruitment of volunteers can also take place through these apps. The usage data collected can be further analyzed in real-time and used to predict the communities that are in dire need of assistance using machine learning or deep learning techniques.

## Conclusions

In this paper, we explored the impact of the COVID-19 pandemic on people globally using social media data. We analyzed over 1 million comments obtained from six social media platforms using a seven-stage context-aware Natural Language Processing (NLP) approach to extract candidate themes or keyphrases which we further categorized into broader themes using thematic analysis. Our results revealed 34 negative themes, out of which 15 are *health-related issues*, *psychosocial issues*, and *social issues* related to the COVID-19 pandemic from the public perspective. Top health-related issues include *increased mortality*, *comparison with other diseases or incidents*, *nature of disease*, and *health concerns*, while top psychosocial issues include *frustrations due to life disruptions*, *panic shopping*, and *expression of fear*. Top social issues include *harassment* and *domestic violence*. Besides the negative themes, 20 positive themes emerged from our results. Some of the positive themes include *public awareness*, *encouragement*, *gratitude*, *cleaner* *environment*, *online learning*, *charity*, *spiritual support*, and *innovative research*. We reflected on our findings and recommend interventions that can help address the health, psychosocial, and social issues based on the positive themes and other remedial ideas rooted in research.

Digital interventions using emerging technologies such as mobile apps, artificial intelligence (AI), internet of things (IoT), and virtual reality will play a major role in delivering remote healthcare (i.e., telemedicine or telehealth) to people in the comfort of their homes, including empowering them to self-manage their health and wellness. This will help to curb the spread of COVID-19 and future infectious diseases since many people will stay away from hospitals (or clinics) to book appointments or see doctors (or other healthcare professionals) unless it is absolutely necessary to visit, thereby keeping health workers and patients safe. These technologies are also useful in providing timely and accurate information about COVID-19 symptoms, diagnosis, treatment, precautionary/safety measures and guidelines, and other relevant information to target audience worldwide. Finally, digital interventions and other interventions discussed in this paper can help address the emotional, physical, and spiritual needs of people who are traumatized or frustrated by the disruptions caused by the pandemic. They also inform governments, health professionals and agencies, and institutions on how to react to the current COVID-19 pandemic, as well as future pandemics.

## Acknowledgements
The authors would like to thank NSERC Discovery Grant for funding this research. They also thank the DeepSense team at Dalhousie University, as well as Compute Canada, for providing the computing infrastructure used to perform our research experiments.

## Authors' Contributions
OO collected data, conducted the experiments, analyzed the results, and wrote the manuscript. CN, DM, BS, and AA collected data and categorized the themes, and reviewed the manuscript. RO is the research supervisor and reviewed the manuscript. FAO is a data analyst and researcher, and reviewed the manuscript. SM is a Psychiatry and Epidemiology researcher, and reviewed the manuscript. CC is a clinical psychologist and reviewed the manuscript.

## Conflicts of Interest
None declared

## Abbreviations
NLP: natural language processing
POS: part of speech